# Transfer-Ensemble Learning based Deep Convolutional Neural Networks for Diabetic Retinopathy Classification

Susmita Ghosh
*Department of Computer Science and Engineering*
*Jadavpur University*
Kolkata, India
susmitaghoshju@gmail.com

Abhiroop Chatterjee
*Department of Computer Science and Engineering*
*Jadavpur University*
Kolkata, India
abhiroopchat1998@gmail.com

*Abstract*—This article aims to classify diabetic retinopathy (DR) disease into five different classes using an ensemble approach based on two popular pre-trained convolutional neural networks: VGG16 and Inception V3. The proposed model aims to leverage the strengths of the two individual nets to enhance the classification performance for diabetic retinopathy. The ensemble model architecture involves freezing a portion of the layers in each pre-trained model to utilize their learned representations effectively. Global average pooling layers are added to transform the output feature maps into fixed-length vectors. These vectors are then concatenated to form a consolidated representation of the input image. The ensemble model is trained using a dataset of diabetic retinopathy images (APTOS), divided into training and validation sets. During the training process, the model learns to classify the retinal images into the corresponding diabetic retinopathy classes. Experimental results on the test set demonstrate the efficacy of the proposed ensemble model for DR classification achieving an accuracy of 96.4%.

## I. INTRODUCTION

Diabetic retinopathy (DR) is a significant contributor to blindness in the working-age population globally. Timely and precise detection of DR is essential for timely intervention and effective disease management. Researchers have employed diverse methodologies, including neural networks, fuzzy sets, and nature-inspired computing, to improve the accuracy of tasks like object tracking, object segmentation, and object detection, making them applicable to computer vision applications [1-4]. Many scholars have made significant contributions to medical image analysis using the above said techniques, aiming to enhance disease diagnosis and detection. In recent years, deep neural nets [5] have shown remarkable advancements in various computer vision tasks, including medical image analysis. This research focuses on classifying diabetic retinopathy into five distinct categories using an ensemble of deep learning models. The ensemble consists of two well-known convolutional neural network (CNN) architectures, namely VGG16 [6] and Inception V3 [7], with a shared fully connected layer for classification. The objective is to leverage the strengths of both models to enhance the accuracy and robustness of diabetic retinopathy classification. For this study, we utilized the pre-trained VGG16 and Inception V3 models, both trained on the ImageNet [8] dataset, as the base models. To prevent overfitting and facilitate transfer learning, we froze a certain depth of layers in each model. The selection of frozen layers was based on a fraction of the total number of layers in each model, allowing them to retain their learned features while adapting to the specific task of diabetic retinopathy classification.

The input images, resized to 224x224 pixels, were fed into the ensemble model, and predictions were generated from both the VGG16 and Inception V3 models. By concatenating the predictions from these models, we aimed to capture a diverse range of features extracted by each network.

Experimental results demonstrate the effectiveness of our proposed ensemble approach for diabetic retinopathy classification, achieving higher accuracy by leveraging the strengths of both VGG16 and InceptionV3 nets.

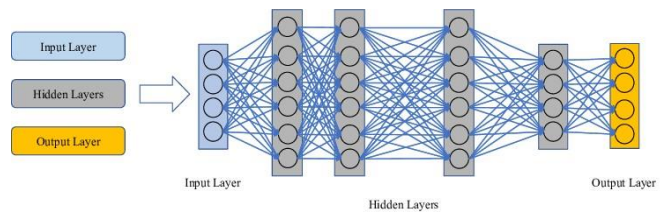

**Fig 1:** General representation of a deep neural network [9].

To evaluate our model, we employed the APTOS 2019 [10] Blindness Detection Challenge dataset, a widely used benchmark dataset in diabetic retinopathy detection. This dataset comprises high-resolution retinal fundus images with corresponding diagnostic labels from expert ophthalmologists. The use of this dataset allowed us to train and evaluate our proposed methodology on diverse and representative samples.

We compared our results in terms of accuracy, precision, recall, and F1-score, providing insights into our newly designed model's performance compared to state-of-the-art techniques. Our findings demonstrate the model's potential for real-world clinical applications.

The rest of this paper is organized as follows: Section 2 presents a review of related works in diabetic retinopathy detection and deep learning. In Section 3, we detail our methodology, including transfer learning with pre-trained VGG16 and Inception V3 models, freezing selected layers, and concatenating predictions. Section 4 covers the experimental setup, including the dataset and evaluation metrics. Next, Section 5 presents the experimental results and their analysis. Finally, Section 6 concludes the paper.

## II. RELATED RESEARCH

Gulshan et al. [11] and Abràmoff et al. [12] proposed a CNN model for automated screening of diabetic retinopathy. They trained deep neural networks using datasets of retinal fundus images. The CNN architecture consisted of multiple convolutional layers followed by max pooling layers to extract features from the images. The extracted features were then passed through fully connected layers for classification. Gargeya and Leng [13] presented a review of deep learning-based approaches for diabetic retinopathy screening. They discussed different CNN architectures used in the literature, including AlexNet, GoogLeNet, and ResNet, for diabetic retinopathy detection. Li et al. [14] proposed a multi-task deep learning framework for simultaneous detection and grading of diabetic retinopathy lesions. Their model effectively detected lesions such as microaneurysms, hemorrhages, and exudates while classifying diabetic retinopathy severity levels. Das et al. [15] proposed a transfer learning approach for diabetic retinopathy detection using CNN models pre-trained on natural image datasets. Their method showed promising results even with a limited number of diabetic retinopathy images. Burlina et al. [16] presented a deep learning model for detecting and classifying retinal lesions related to diabetic retinopathy. Their system incorporated multiple CNN architectures to handle different lesion types, improving overall performance. Rajalingappaa et al. [17] introduced a deep learning model for diabetic retinopathy classification using retinal fundus images. Their approach utilized a combination of CNN and Long Short-Term Memory (LSTM) networks to capture spatial and temporal patterns.

This research focuses on diabetic retinopathy classification using an ensemble of deep learning models, VGG16 and Inception V3, with a shared fully connected layer. Both models are pretrained on ImageNet, and specific layers were frozen to prevent overfitting and enable transfer learning. By leveraging the strengths of both architectures, and combined feature extraction aims to enhance accuracy in classifying diabetic retinopathy into five distinct categories.

## III. METHODOLOGY

In this work, we propose a transfer-ensemble based method for diabetic retinopathy classification. Our approach utilizes the power of pre-trained convolutional neural network (CNN) models, specifically VGG16 and Inception V3, to extract high-level features from retinal images. The methodology consists of several key steps to enhance the performance. The block diagram of the proposed method has been demonstrated in Fig. 2.

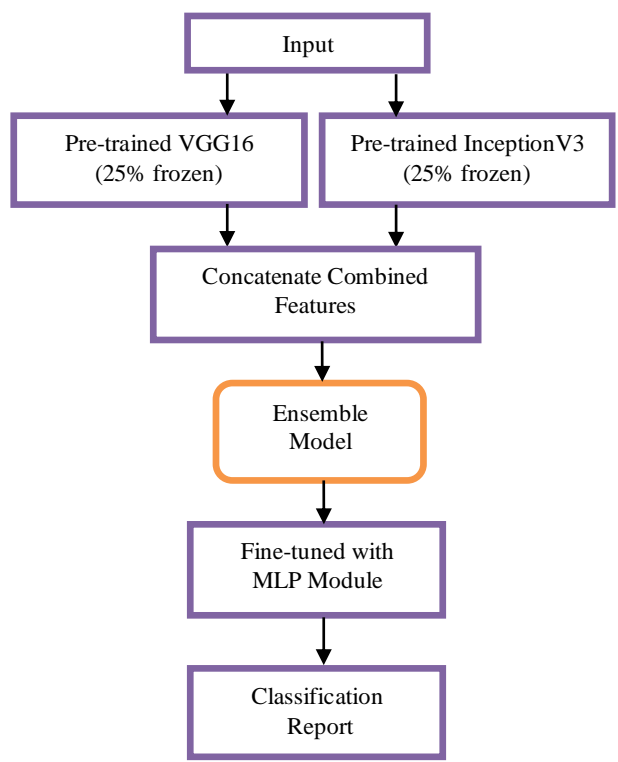

**Fig 2:** Flowchart of the proposed model

**(A) Transfer Learning:**
We utilize the pre-trained VGG16 and Inception V3 models, which have been trained on large-scale datasets, like the ImageNet. By utilizing these models, we can benefit from their learned feature representations that capture various visual patterns and structures and also require less computational power.

**(B) Model Freezing:**
To avoid overfitting and allow effective transfer learning, we chose to freeze a substantial portion of each pre-trained model. Approximately 25% of the layers in VGG16 and Inception V3 were frozen, ensuring that the initial representations learned on ImageNet remain intact and generalizable to our diabetic retinopathy classification task.

**(C) Ensemble Learning:**
As mentioned, our methodology extracts the strengths of both the neural nets through ensemble learning. We obtained predictions from the frozen VGG16 and Inception V3 models for each input image and concatenated them. This enabled the ensemble model to capture diverse and complementary feature representations, leading to improved classification performance.

**(D) Architecture of the Model and Training Strategy:**
To enhance the classification performance, we introduced several key components in our methodology. First, after obtaining predictions from the frozen VGG16 and Inception V3 models, we concatenated them to capture diverse and complementary feature representations. Next, we applied Global Average Pooling to ensure compatibility between the models' outputs for seamless merging. Subsequently, a fully connected layer with 256 neurons and ReLU activation facilitated learning higher-level features and complex patterns. To prevent overfitting, we incorporated dropout regularization in the fully connected layer. Finally, the output layer consisted of a softmax activation function with five neurons, enabling confident and accurate predictions for the five classes of diabetic retinopathy severity levels. Through these strategic additions, our ensemble model demonstrates improved classification.

## IV. EXPERIMENTAL SETUP

As stated, experiment with the proposed neural net model is conducted using APTOS dataset. Details of this experimental setup have been described below in detail.

**(A) Dataset Used:**
The APTOS dataset consists of a large collection of high-resolution retinal fundus images along with corresponding diagnostic labels provided by expert ophthalmologists. A total of 6034 images were taken from 5 different classes. Table 1 shows the number of images used for different categories.

**(B) Image Preprocessing:**
Each image is resized to a dimension of 224x224 pixels and normalized by dividing the pixel values by 255. There are five classes in the image dataset: Mild DR, Moderate DR, No DR, Proliferate DR, and Severe DR. A sample image from each of the classes has been shown in Fig. 3.

**Table 1:** Images taken from APTOS 2019 dataset

| Types of Classes | Number of Images |
|---|---|
| Mild DR | 1624 |
| Moderate DR | 999 |
| No DR | 1805 |
| Proliferate DR | 772 |
| Severe DR | 834 |
| **Total Images** | **6034** |

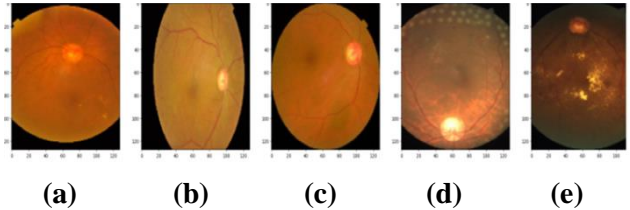

(a)    (b)    (c)    (d)    (e)

**Fig. 3:** Images taken from APTOS dataset. (a) Mild DR, (b) Moderate DR, (c) No DR, (d) Proliferate DR, and (e) Severe DR

We have also considered other performance indices e.g., precision, recall, F1-score, Top-1% error for comparison purposes. Confusion matrix is also considered.

**(D) Parameters Taken:**
Table 2 shows the parameters and their corresponding values used for our experimentation.

**TABLE 2:** Experimental setup

| Parameters | Values |
|---|---|
| Learning Rate | 0.0001 |
| Batch Size | 16 |
| Max Epochs | 40 |
| Optimizer | Adam |
| Loss Function | Categorical Cross-entropy |

**(E) Model Training:**
The dataset is split into training and test sets. The split is performed with a test size of 20% and stratified sampling to maintain class balance.

## V. RESULTS

To evaluate the effectiveness of the proposed methodology, five different performance measuring criteria are used. These are: accuracy, precision, recall, and F1-score. The corresponding results are put in Table 3. Results are promising in nature in terms of various performance indices, yielding 96.4% accuracy for the ensemble model.

**TABLE 3:** Performance metrics on the APTOS dataset.

| Metrics Used | Ensemble (rounded) |
|---|---|
| Precision | 0.96 |
| Recall | 0.96 |
| F1-score | 0.96 |
| Accuracy (%) | 96 |
| Top-1 error (%) | 4.0 |

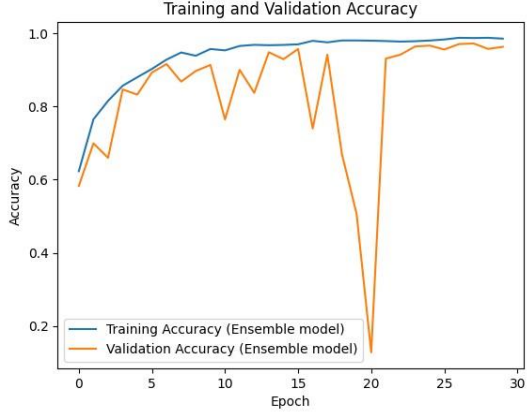

**Fig 4:** Variation of training and validation accuracy with epochs of the fine-tuned ensemble model.

From Fig. 4 it is seen that both training and validation accuracy increases over epochs. At first we notice a steady increase in accuracy for both the validation and training sets. This suggests that the model is effectively learning from the data and making accurate predictions. Towards the end of training, the accuracies stabilize towards higher value. Since the pre-trained models were used, the weights were already updated toward the optimum values and hence the loss (both training and validation) are minimal from the very beginning, as seen in Fig. 5. This indicates that the model is becoming proficient at generalizing patterns and minimizing errors.

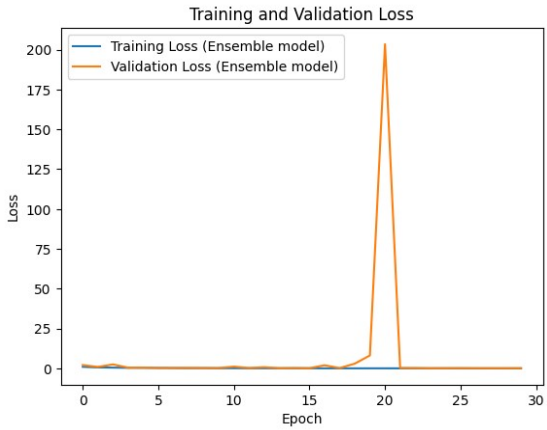

**Fig 5:** Variation of training and validation loss with epochs of the fine-tuned ensemble model.

Each of the CNNs was trained and tested on the APTOS dataset, and the simulations were repeated 20 times to account for variations in the results. For each simulation, we randomly split the dataset into training and testing sets, maintaining the same split ratio for consistency across all simulations. After each simulation, we recorded the evaluation scores.

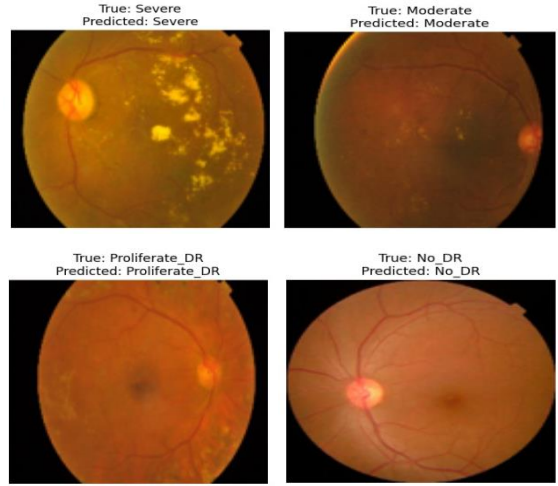

**Fig 6:** Class predictions on unseen images by the proposed ensemble method.

The accuracy values obtained using the two individual deep neural networks (VGG16 and Inception V3) and that for the proposed ensemble model are put in Table 4. Here the average accuracy values over 20 simulations are given. This result confirms the effectiveness of the ensemble approach as compared to the individual net.

**Table 4:** Comparison of our proposed ensemble model with individual models.

| Models | Accuracy (%) |
| --- | --- |
| VGG16 | 92.6 |
| Inception V3 | 94.7 |
| Ensemble | **96.4** |

For visual illustrations, Fig. 6 shows the predictions made by our fine-tuned ensemble model on some unseen images of APTOS, 2019 dataset.

**TABLE 5:** Classification performance of the proposed method as to the state-of-the-art models for the APTOS dataset.

| Methods | Accuracy (%) |
| --- | --- |
| (Dondeti et al., 2020) [18] | 77.90 |
| (Bodapati et al., 2020) [19] | 81.70 |
| (Liu et al., 2020) [20] | 86.34 |
| (Kassani et al., 2019) [21] | 83.09 |
| (Bodapati et al., 2021) [22] | 82.54 |
| (Sikder et al., 2021) [23] | 94.20 |
| (Alyoubi et al., 2021) [24] | 89.00 |
| Proposed Framework | **96.40** |

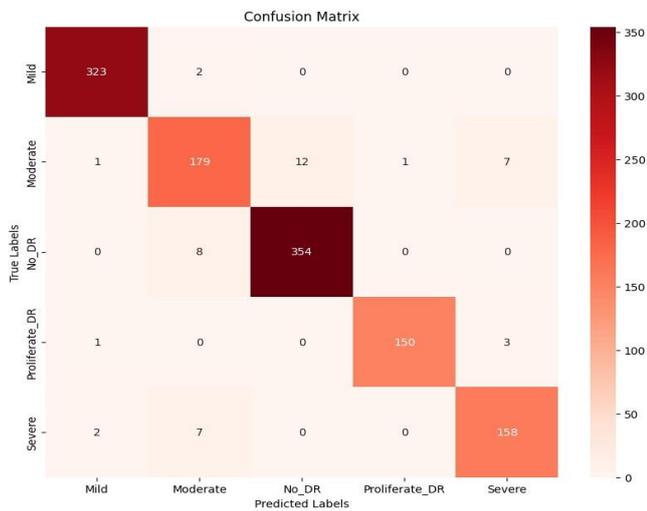

**Fig 7:** Confusion matrix for five different stages of DR for the proposed ensemble method.

Performance of the proposed ensemble model is also compared with seven other state-of-the-art models and the results are shown in Table 5. These results also establish the superiority of the proposed ensemble model.

Finally, the confusion matrix obtained through experimentation is also shown in Fig. 7. This matrix also corroborates our earlier findings regarding the efficacy of the proposed approach.

## VI. CONCLUSION

In conclusion, the present work proposes a transfer-ensemble based approach for enhanced diabetic retinopathy classification. By combining the complementary strengths of VGG16 and Inception V3 models, we achieved improved classification performance compared to individual nets. Our ensemble method, which concatenates the feature maps from VGG16 and Inception V3, demonstrated enhanced discriminative power in capturing diverse image patterns. Further research can explore combinations of other models and investigate ensemble methods' generalizability across different retinal diseases and imaging modalities. Overall, our study justifies the importance of ensemble learning in enhancing diagnostic accuracy. Moreover, the proposed transfer-ensemble approach holds promising potential to contribute to the development of computer-aided diagnosis systems for diabetic retinopathy diagnosis.

## Acknowledgement

A part of this work has been supported by the IDEAS - Institute of Data Engineering, Analytics and Science Foundation, The Technology Innovation Hub at the Indian Statistical Institute, Kolkata through sanctioning a Project No /ISI/TIH/2022/55/ dtd. September 13, 2022.

## References

[1] Haykin, S. (2008). Neural Networks and Learning Machines. Pearson.

[2] Mondal, A., Ghosh, S., & Ghosh, A. (2017). Partially camouflaged object tracking using modified probabilistic neural network and fuzzy energy based active contour. International Journal of Computer Vision, 122, 116-148.

[3] Ross, T. J. (2004). Fuzzy Logic with Engineering Applications. Wiley.

[4] Dehuri, S., Ghosh, S., & Cho, S. (2011). Integration of swarm intelligence and artificial neural network. World Scientific, 78.

[5] Goodfellow, I., Bengio, Y., & Courville, A. (2016). Deep Learning. MIT Press.

[6] Simonyan, K., & Zisserman, A. (2014). Very Deep Convolutional Networks for Large-Scale Image Recognition. arXiv preprint arXiv:1409.1556.

[7] Szegedy, C., Vanhoucke, V., Ioffe, S., Shlens, J., & Wojna, Z. (2016). Rethinking the Inception Architecture for Computer Vision. In Proceedings of the IEEE Conference on Computer Vision and Pattern Recognition (CVPR) (pp. 2818-2826). doi:10.1109/CVPR.2016.308.

[8] Deng, J., Dong, W., Socher, R., Li, L.-J., Li, K., & Fei-Fei, L. (2009). ImageNet: A Large-Scale Hierarchical Image Database. In 2009 IEEE Conference on Computer Vision and Pattern Recognition (CVPR) (pp. 248-255). IEEE. doi:10.1109/CVPR.2009.5206848.

[9] URL: https://www.researchgate.net/figure/General-Deep-Learning-Neural-Network_fig1_336179118.

[10] Mello, V., Silva, G. F., Mendonça, J. S., & Lins, R. D. (2019). APTOS 2019 Blindness Detection. Kaggle. Available from: https://www.kaggle.com/c/aptos2019-blindness-detection.

[11] Gulshan, V., et al. (2016). "Development and Validation of a Deep Learning Algorithm for Detection of Diabetic Retinopathy in Retinal Fundus Photographs." JAMA, 316(22), 2402-2410. doi:10.1001/jama.2016.17216.

[12] Abràmoff, M. D., et al. (2018). "Improved Automated Detection of Diabetic Retinopathy on a Publicly Available Adjudicated Database." Investigative Ophthalmology & Visual Science, 59(1), 314-321. doi:10.1167/iovs.17-22650.

[13] Gargeya, R., & Leng, T. (2017). "Automated Identification of Diabetic Retinopathy Using Deep Learning." Ophthalmology, 124(7), 962-969. doi:10.1016/j.ophtha.2017.02.008.

[14] Li, Z., et al. (2017). "Multimodal Retinal Image Analysis via Deep Learning for the Detection of Diabetic Retinopathy." IEEE Journal of Biomedical and Health Informatics, 21(2), 411-420. doi:10.1109/JBHI.2016.2636365.

[15] Das, A., et al. (2018). "A Deep Learning Based Approach for Automated Detection and Grading of Diabetic Retinopathy Using Retinal Fundus Images." Computers in Biology and Medicine, 101, 64-77. doi:10.1016/j.compbiomed.2018.07.021.

[16] Burlina, P. M., et al. (2017). "Automated Grading of Age-Related Macular Degeneration From Color Fundus Images Using Deep Convolutional Neural Networks." JAMA Ophthalmology, 135(11), 1170-1176. doi:10.1001/jamaophthalmol.2017.3782.

[17] Rajalingappaa, M., et al. (2017). "Deep Convolution Neural Network-Based Diabetic Retinopathy Screening Model." Journal of Healthcare Engineering, 2017, 6586596. doi:10.1155/2017/6586596.

[18] Dondeti, V., Bodapati, J.D., Shareef, S.N., & Veeranjaneyulu, N. (2020). Deep convolution features in non-linear embedding space for fundus image classification. Revue d'Intelligence Artificielle, 34(3), 307-313.

[19] Bodapati, J.D., Naralasetti, V., Shareef, S.N., Hakak, S., Bilal, M., Maddikunta, P.K.R., & Jo, O. (2020). Blended multi-modal deep convnet features for diabetic retinopathy severity prediction. Electronics, 9(6), 914.

[20] Liu, H., Yue, K., Cheng, S., Pan, C., Sun, J., & Li, W. (2020). Hybrid model structure for diabetic retinopathy classification. Journal of Healthcare Engineering, 2020.

[21] Kassani, S.H., Kassani, P.H., Khazaeinezhad, R., Wesolowski, M.J., Schneider, K.A., & Deters, R. (2019). Diabetic retinopathy classification using a modified Xception architecture. In 2019 IEEE International Symposium on Signal Processing and Information Technology (ISSPIT) (pp. 1-6). IEEE.

[22] Bodapati, J.D., Shaik, N.S., & Naralasetti, V. (2021). Composite deep neural network with gated-attention mechanism for diabetic retinopathy severity classification. Journal of Ambient Intelligence and Humanized Computing, 1-15.
[23] Sikder, N., Masud, M., Bairagi, A.K., Arif, A.S.M., Nahid, A.A., & Alhumyani, H.A. (2021). Severity classification of diabetic retinopathy using an ensemble learning algorithm through analyzing retinal images. Symmetry, 13(4), 670.
[24] Alyoubi, W.L., Abulkhair, M.F., & Shalash, W.M. (2021). Diabetic retinopathy fundus image classification and lesions localization system using deep learning. Sensors, 21(11), 3704.